\documentclass[sigplan]{acmart}

\AtBeginDocument{%
  \providecommand\BibTeX{{%
    \normalfont B\kern-0.5em{\scshape i\kern-0.25em b}\kern-0.8em\TeX}}}

\setcopyright{acmcopyright}

\copyrightyear{2021}
\acmYear{2021}
\setcopyright{acmlicensed}\acmConference[MM '21]{Proceedings of the 29th ACM International Conference on Multimedia}{October 20--24, 2021}{Virtual Event, China}
\acmBooktitle{Proceedings of the 29th ACM International Conference on Multimedia (MM '21), October 20--24, 2021, Virtual Event, China}
\acmPrice{15.00}
\acmDOI{10.1145/3474085.3475242}
\acmISBN{978-1-4503-8651-7/21/10}

\def\eg{\emph{e.g.}} 
\def\ie{\emph{i.e.}} 
\def\etal{\emph{et al.}}
\def\etc{\emph{etc.}}
\usepackage{subfigure}
\usepackage{bm}
\usepackage{multirow}
\renewcommand\footnotetextcopyrightpermission[1]{} 
\begin{document}
\fancyhead{}
\title{DC-GNet: Deep Mesh Relation Capturing Graph Convolution Network for 3D Human Shape Reconstruction}

\author{Shihao Zhou$^{1\#}$, Mengxi Jiang$^{1\#}$, Shanshan Cai$^1$, Yunqi Lei$^1$}
\authornote{Corresponding Author.\\$^\#$Both authours contributed equally to this research.}
\affiliation{%
  \institution{$^1$Department of Computer Science, School of Informatics, Xiamen University, 361005, }
  \city{Xiamen}
  \state{Fujian Province}
  \country{China}
}
\email{{shzhou,jiangmengxi,sscai}@stu.xmu.edu.cn,yqlei@xmu.edu.cn}


\renewcommand{\shortauthors}{Trovato and Tobin, et al.}

\begin{abstract}
In this paper, we aim to reconstruct a full 3D human shape from a single image. Previous vertex-level and parameter regression approaches reconstruct 3D human shape based on a pre-defined adjacency matrix to encode positive relations between nodes. The deep topological relations for the surface of the 3D human body are not carefully exploited. Moreover, the performance of most existing approaches often suffer from domain gap when handling more occlusion cases in real-world scenes.

In this work, we propose a \textbf{D}eep Mesh Relation \textbf{C}apturing \textbf{G}raph Convolution \textbf{Net}work, DC-GNet, with a shape completion task for 3D human shape reconstruction. Firstly, we propose to capture deep relations within mesh vertices, where an adaptive matrix encoding both positive and negative relations is introduced. Secondly, we propose a shape completion task to learn prior about various kinds of occlusion cases. Our approach encodes mesh structure from more subtle relations between nodes in a more distant region. Furthermore, our shape completion module alleviates the performance degradation issue in the outdoor scene. Extensive experiments on several benchmarks show that our approach outperforms the previous 3D human pose and shape estimation approaches.
\end{abstract}



\keywords{3D human shape reconstruction, Graph Convolution Network, Deep mesh relation capturing}


\maketitle

\section{Introduction}
\label{intro}
3D human pose and shape estimation is a fundamental yet challenging task in computer vision. There are plenty of approaches proposed to accurately capture 2D pose and even 3D joint locations~\cite{SunXLW19,zhaoCVPR19semantic,Sharma_2019_ICCV,fabbri2020compressed,Zhou_2019_ICCV,Li_2019_CVPR}. Since sparse joints alone cannot provide enough information for analyzing humans~\cite{kolotouros2019convolutional}, incremental recent works interest in recovering the 3D mesh of a human body, where the 3D surface is defined. 

\begin{figure}[]
\centering
\includegraphics[width=\linewidth]{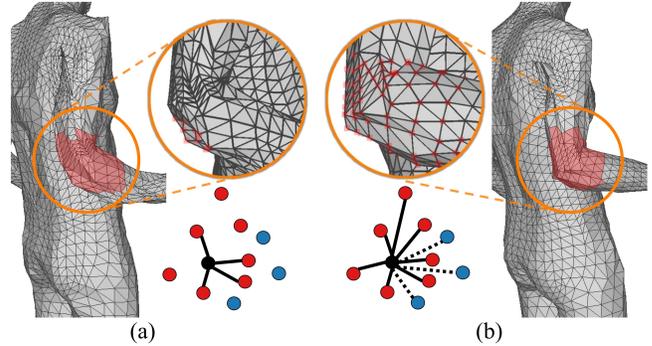}
\caption{Illustration of different strategies to reason local structure. (a) Previous popular approaches are based on a pre-defined adjacency matrix that encodes only positive relations with physically connected nodes. (b) Our approach learns deep relations~(\ie, positive and negative) between nodes in a more distant region. The inference node, positively related node and negatively related node are shown in the black, red and blue circle, respectively. The solid line denotes a positive relationship, while the dashed line denotes a negative relationship.}
\label{pic:mot}
\end{figure}

To obtain 3D mesh for a human being in an image, optimization-based approaches generate a reliable human body fitting~\cite{bogo2016keep,lassner2017unite}. Unfortunately, their slow inference speed and sensitivity to initialization have shifted the focus to regression-based approaches, which directly regresses mesh coordinates~\cite{kolotouros2019convolutional,Choi_2020_ECCV_Pose2Mesh,pointMesh} or the parameters~\cite{kanazawa2018end-to-end,omran2018neural,pavlakos2018learning} of the human body model~(\eg, SCAPE~\cite{SCAPE2005} and SMPL(-X)~\cite{smpl2015,smplxCvpr2019,MANO2017}). Although regression-based methods achieve clear performance improvement in constrained environments, there are still limitations hinders better performance under the real scenario.

\begin{figure}[htbp]
\centering
\subfigure[]{
    \begin{minipage}[t]{0.5\linewidth}
    \label{pic:adjCompA}
        \centering
        \includegraphics[width=\linewidth]{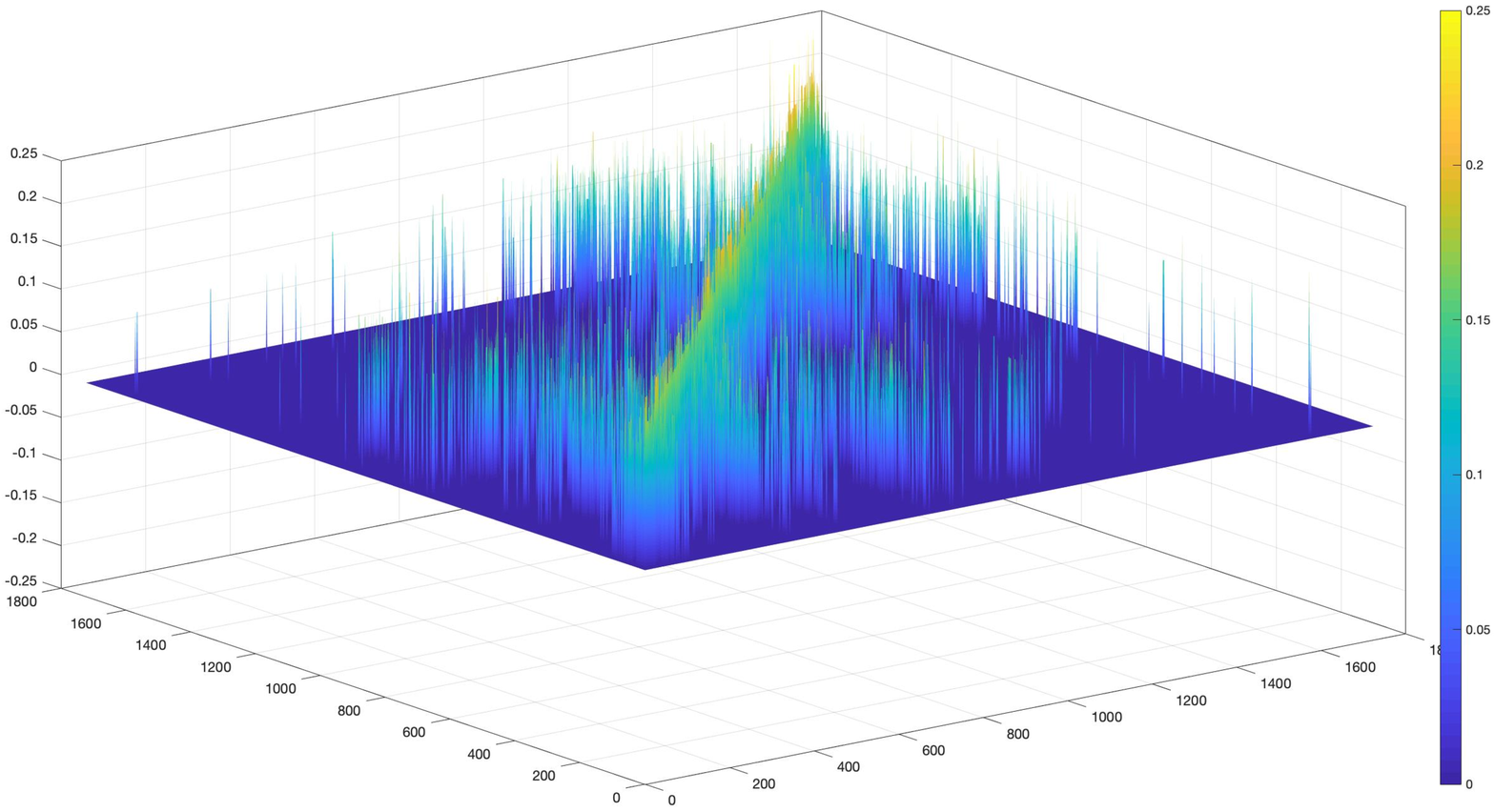}\\
    \end{minipage}%
}%
\subfigure[]{
    \begin{minipage}[t]{0.5\linewidth}
    \label{pic:adjCompB}
        \centering
        \includegraphics[width=\linewidth]{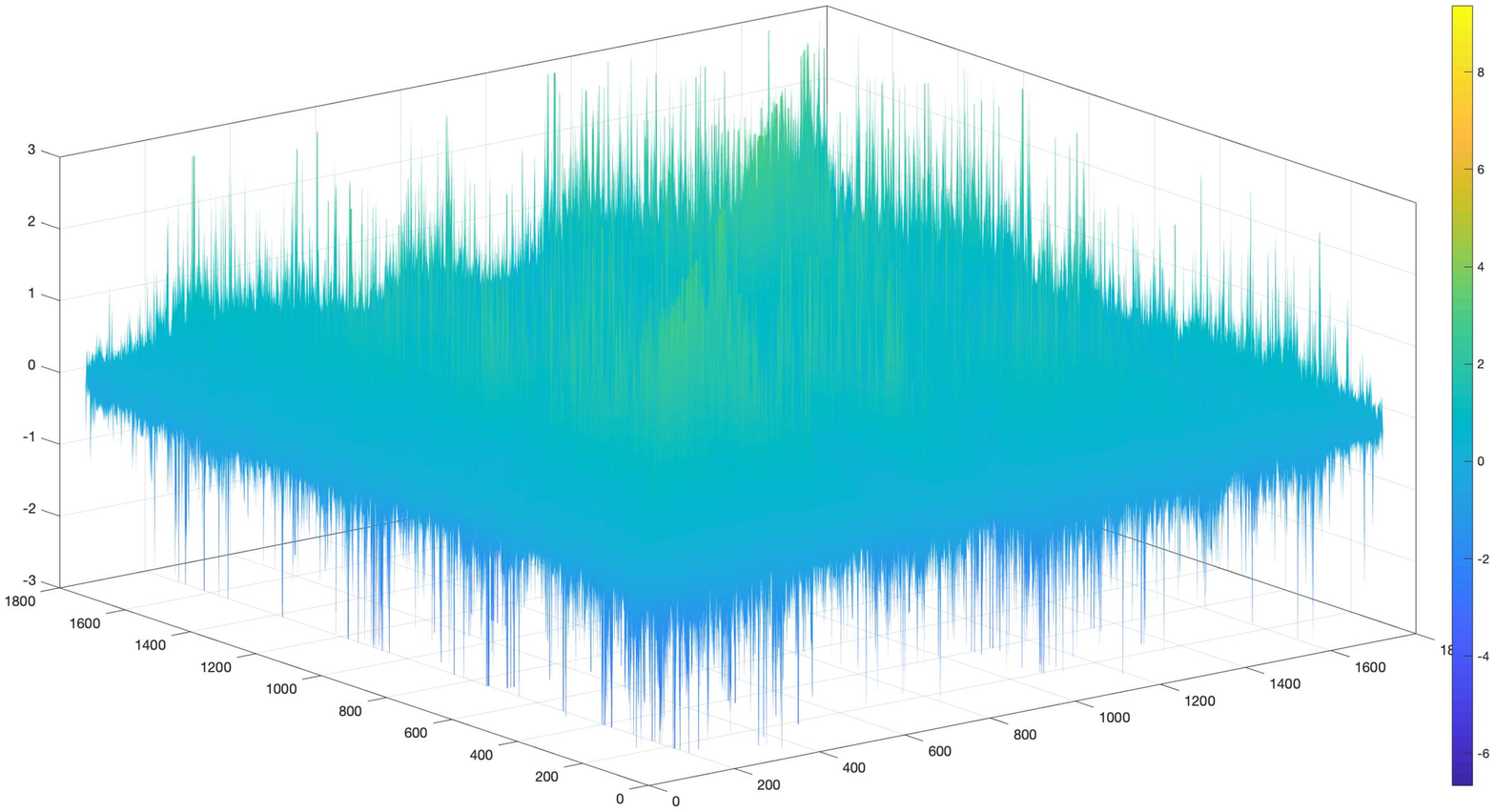}\\
    \end{minipage}%
}%
\centering
\caption{Visualization of the different adjacency matrix. (a) A pre-defined adjacency matrix with encoding only positive relations between physically connected nodes. (b) Our adjacency matrix learns subtle relations~(\ie, positive and negative) between nodes in a more distant region.}
\label{fig:compare_fig}
\end{figure}
Firstly, 
most of existing regression-based approaches including vertex-level regression approaches~\cite{kolotouros2019convolutional,Choi_2020_ECCV_Pose2Mesh} and parameter regression approaches~\cite{kanazawa2018end-to-end,pavlakos2018learning} are based on a fixed adjacency matrix to encode the inherent shape nodes relations, which ignores deep relations between shape nodes and focus only on physically connected nodes, as shown in Figure~\ref{pic:mot}(a). As a result, models often can not fully explore the spatial relations within the human body, which owns a highly related structure. 

Previous works~\cite{zhaoCVPR19semantic,Doosti_2020_CVPR} attempt to explore a joint-to-joint topological structure for skeleton joints in the 3D pose estimation task. However, compared to the 3D pose estimation task with skeleton representation, a full 3D human shape reconstruction task needs to infer node-to-surface relations rather than a simple joint-to-joint. Thus, these approaches cannot be applied directly to the 3D shape estimation. To the best of our knowledge, exploiting the deep topological relations for the surface of the 3D human body is an unexplored yet important problem to the current 3D human shape reconstruction approaches. 

Secondly, the full human body information is usually difficult to obtain in real scenarios due to various occlusions. Existing 3D annotation datasets that are collected in the limited indoor environment not carefully simulate such cases of body partially missing, which forms a gap issue of appearance domain~(\ie,~the performance degradation in real-world scenes). Therefore, the generalization of existing shape reconstruction approaches trained on indoor 3D data are often poor when handling with more occlusion cases in real-world scenes, resulting in the performance degradation.

In this paper, to alleviate the above issues, we propose a Deep Mesh Relation Capturing Graph Convolution Network, namely DC-GNet, with a shape completion task. Firstly, to capture deep relation among mesh vertexes of body shape, we impose an adaptive adjacency matrix to learn both positive and negative relationships between nodes in a more distant region, as shown in Figure~\ref{pic:adjCompB}. Base on this learnable matrix, our network can aggregate subtle information from not only physically connected nodes but also nodes with long-distance, as shown in Figure~\ref{pic:mot}(b). 

\begin{figure*}[]
\centering
\includegraphics[width=\linewidth]{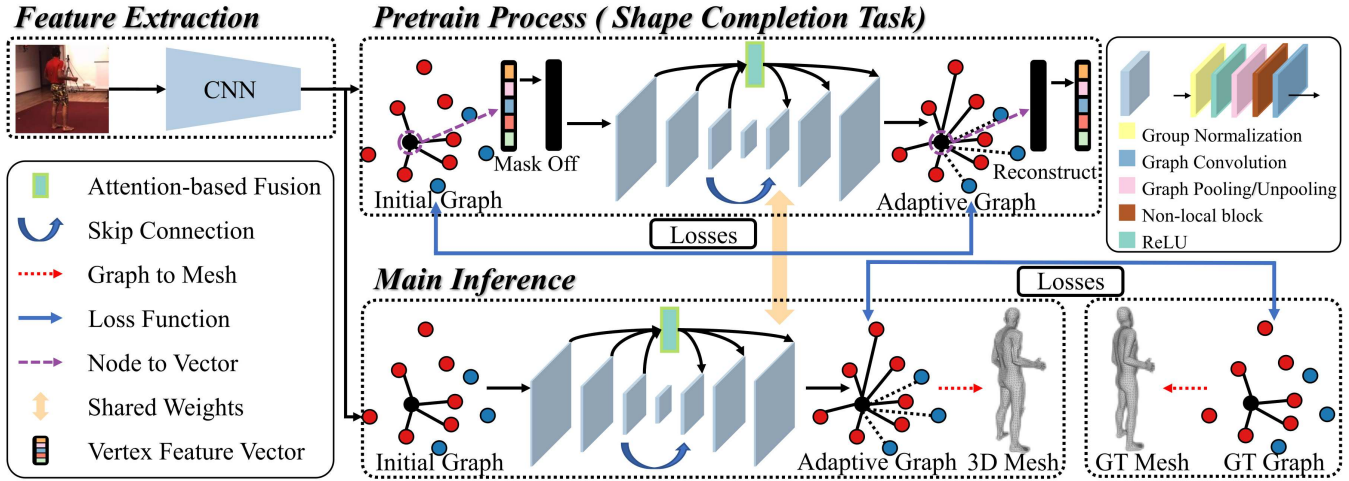}
\caption{Overview of DC-GNet. The workflow contains three parts, a feature extraction stage to generate the initial graph from a single image, a pretrain process to learn an adaptive graph with a shape completion task and the main inference phase to reconstruct the 3D mesh.  }
\label{pic:overview}
\end{figure*}

Secondly, we propose a shape completion task to alleviate the gap issue of appearance domain. Specifically, we first fabricate artificial holes on the surface of the training body shape data. Then, we force the network to recover a full body shape to learn prior about various kinds of human body part missing case. The overview of our approach is shown in Figure~\ref{pic:overview}. Concretely, our network is started with the initial estimation, which is the feature extraction stage. In the pretrain process, the image features are inputted into DC-GNet with the proposed shape completion task, where the network learns the prior for occlusion cases. This network is further trained in the main inference process for the 3D human shape estimation.

Extensive evaluated experiments are conducted on several public benchmarks~\cite{2014Human3,mono-3dhp2017}, and the results show that our proposed approach outperforms the previous state-of-the-art 3D pose and shape estimation methods. Moreover, qualitative results on in-the-wild datasets~\cite{2014Human,lin2014microsoft} show that our approach has learned better representation and generalization ability through making better use of the topology structure of the human body.

We summarize our contributions as follows:
\begin{itemize}
\item[$\bullet$] We propose a Deep Mesh Relation Capturing Graph Convolution Network, DC-GNet, to reconstruct 3D human shape from a single RGB image. It is the first attempt to learn deep relations between nodes among human mesh vertices and consider reasoning from more than partial structure, which boosts the model capturing complex local deformation.

\item[$\bullet$] We first propose a shape completion module as an auxiliary task to alleviate the appearance domain gap issue between indoor and outdoor scenes, and thus our model can benefit from the rich indoor data source to learn the inference paradigm that can be well applied in natural environments.
\item[$\bullet$] Extensive experimental results across several benchmarks demonstrate the effectiveness of exploring deep relations among mesh vertices, through comparing with the previous state-of-the-art 3D human shape reconstruction methods.
\end{itemize}

\section{Related Work}
Although numerous approaches have been proposed to boost the topic of 3D pose estimation in the form of a skeleton in the last few years~\cite{20163D,7298976,BMVC.28.80,Zhou2019MonoCap,Suman2013A,towards3D_shhzhou}, we will focus on closely-related works reconstructing the whole shape and pose in this Section~\cite{Guler_2019_CVPR,Sun_2019_ICCV,kolotouros2019spin}.
\subsection{3D Human Pose and Shape Estimation}
\label{relate:3d}
Instead of a skeleton, recovering the shape of human body is a more challenging task. Bogo~\etal~\cite{bogo2016keep} firstly introduced the fully automatic model-based approach, SMPLify, to estimate 3D human shape and pose from 2D pose by fitting a classical human body model SMPL. After that, Lassner~\etal~\cite{lassner2017unite} applied SMPLify for building a dataset with fairly successful 3D fits. Beyond SMPLify, many different model-based approaches were proposed to explore including adversarial prior~\cite{kanazawa2018end-to-end,kocabas2019vibe}, temporal information~\cite{arnab2019exploiting,KanazawaVideo}, or even dealing with multiple humans~\cite{Zanfir_2018_CVPR,jiang2020coherent}. More recently, instead of predicting parameters of the model, model-free methods that directly regress each vertex were proposed to avoid representation issues~\cite{3dbaseline,pavlakos2018learning}. Venkat~\etal~\cite{Venkat_2019_ICCV} captured the local deformation by learning the ``implicitly structure" of human mesh. Similarly, Kolotouros~\etal~\cite{kolotouros2019convolutional} directly regressed the vertices of a template mesh to explore the topological structure explicitly. Moon and Lee~\cite{Moon_2020_ECCV_I2L-MeshNet} proposed an image-to-lixel network to model the prediction uncertainty for each mesh vertex. More interesting, many works~\cite{Zhu_2019_CVPR,bhatnagar2020ipnet,Habermann_2020_CVPR} tried to obtain clothed mesh.

\subsection{Graph Convolution Network for 3D shape Reconstruction}
\label{relate:gcn}
Recently, there has been a surge of approaches considering Graph Convolution Network(GCN) for capturing the graph structure of mesh, due to the graph-like nature of human mesh. Choi~\etal~\cite{Choi_2020_ECCV_Pose2Mesh} recovered 3D mesh from the 2D input with GCN in a coarse-to-fine fashion. Kolotouros~\etal~\cite{kolotouros2019convolutional} explored mesh structure and leveraged spatial locality via applying GCN to directly regress the vertices of the SMPL model. Hugo~\etal~\cite{cloth3d} proposed to generate 3D clothed human with graph convolution variational Auto-Encoder~(GCVAE). Simultaneously, some works~\cite{Venkat_2019_ICCV,GabeurFMSR2019} intended to deal with mesh vertices as point clouds for capturing deformation.

Similarly, we also adopt GCN to process the mesh structure. Different from previous approaches, instead of simply applying convolution operation to aggregate information from direct neighbors, we incorporate an adaptive adjacent matrix to obtain local structure from both physically connected nodes and distant ones with deep relations.

\subsection{Relations Capture via Learnable Adjacency Matrix}
\label{relate:a}
The adjacency matrix is a necessary component in GCN. In the 3D shape Reconstruction task, existing regression-based approaches usually use a pre-defined adjacency matrix, in which only positive relations between physically connected nodes are encoded. Such a pre-defined adjacency matrix is not able to capture complex local surface deformation, as shown in Figure~\ref{pic:mot}(a). Replacing a pre-defined adjacency matrix with learnable ones is a common strategy in the other GCN-based computer vision applications, such as skeleton estimation~\cite{zhaoCVPR19semantic,Doosti_2020_CVPR} or action recognition task~\cite{adpAR_cvpr19,aaai_18_adp}. These works use different updating strategies to learn adjacency matrix for different purposes. Zhao \etal~\cite{zhaoCVPR19semantic} learns adjacent matrix for describing subtle semantic relations within human skeleton joints. Doosti~\etal~\cite{Doosti_2020_CVPR} keeps the connectivity of the graph structure in joint hand and object pose estimation task. Against the action recognition task, Yan~\etal~\cite{aaai_18_adp} captures the body skeletons dynamics information to meet the specific demands in skeleton modeling. Shi~\etal~\cite{adpAR_cvpr19} learns the topology structure of the graph and skeleton samples for the flexibility of model.

These strategies only focus on joint-to-joint relations, which can not capture the node-to-surface relations existing in a full 3D human shape reconstruction task. In this paper, against the 3D shape reconstruction, we first propose a novel updating strategy for the learnable adjacency matrix to explore node-to-surface relations.

\begin{figure*}[]
\centering
\includegraphics[width=\linewidth]{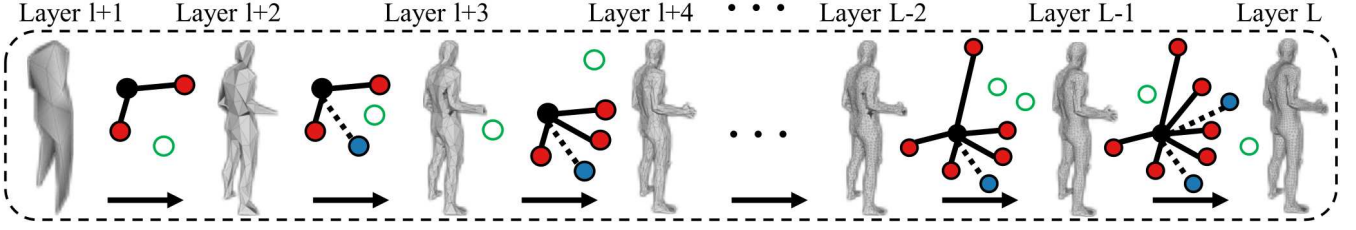}
\caption{Visualization of the mesh from different layers in the decoder part of the U-Net. The refining process generates the final prediction from a coarse graph by adding nodes~(the green hollow circle).}
\label{pic:Process}
\end{figure*} 

\section{Problem Setup}
\label{sec:problemSetup}
Our input is a cropped image, which is centered around a person. For each input image, an image-based convolutional network is applied as a feature extractor and outputs a 2048-D feature vector for every single vertex in the graph.

Our network is started with the initial estimation, which is the feature extraction stage shown in the Figure~\ref{pic:overview}.

Previous approaches had already adopted GCN to process graph-like human mesh. The network is composed of basic graph convolution operations~\cite{kipf2017semi-supervised}, which is defined as:
\begin{flalign}
\bm{X}_{out} = \sigma(\bm{AX}_{in}\bm{W}),
\label{eq:gcn}
\end{flalign}
where $\bm{A} \in \mathbb{R}^{N \times N}$ is a pre-defined adjacency matrix of the graph, $\bm{X}_{in} = \{x_i\}_{i=1}^N \in \mathbb{R}^{N \times k}$ is the input feature matrix, $\bm{W} \in \mathbb{R}^{k \times h}$ is the trainable weight matrix, $\bm{X}_{out} \in \mathbb{R}^{N \times h}$ is the output feature matrix, and $\sigma$ is the activation function. Specifically, $N$ is nodes of the input graph, $k$ and $h$ are the input features and output features for each node, separately. 

The graph convolution described in Eq.~\eqref{eq:gcn} is calculated based on a pre-defined adjacency matrix, which only encodes positive relations between physically connected nodes. As a result, the complex local structure can not be carefully captured.

\section{Proposed Approach}
In this Section, we present our approach. First, in Subsection~\ref{subsec:local}, we describe the proposed network for obtaining effectively local structure information.

Next, in Subsection~\ref{subsec:shape}, we describe our proposed shape completion task that alleviates the gap issue of appearance domain. Finally, Subsection~\ref{app:loss} present the loss functions that we used for training.

\subsection{Relations Capture for the 3D Shape Reconstruction}
\label{subsec:local}
Instead of using a pre-defined adjacency matrix, we propose to use an adaptive adjacency matrix to learn subtle relationships between nonadjacent nodes.  
To achieve this, we rewrite Eq.~\eqref{eq:gcn} as:
\begin{flalign}
\bm{X}_{out} = \sigma(\hat{\bm{A}}\bm{X}_{in}\bm{W}).
\label{eq:Ngcn}
\end{flalign}
where $\hat{\bm{A}}=\bm{A}+\bm{I}$ is a learnable adjacency matrix. $\bm{I}$ is the identity matrix. Based on Eq.~\eqref{eq:Ngcn}, our network is able to infer local structure from nodes with subtle relations in a more distant region. Specifically, not only relations between nodes belong to the same semantic part~(\eg, nodes on arm and elbow that belong to the same limb can be leveraged for inference), but nodes with far distance~(\eg, one node on the left elbow can be related to the node on the right elbow) are encoded into the network. Moreover, following \cite{stGCN_iccv19}, we introduce a non-local block ~\cite{Wang_2018_CVPR} to facilitate a holistic processing of the full body.

\begin{figure}[!t]
\centering
\includegraphics[width=\linewidth]{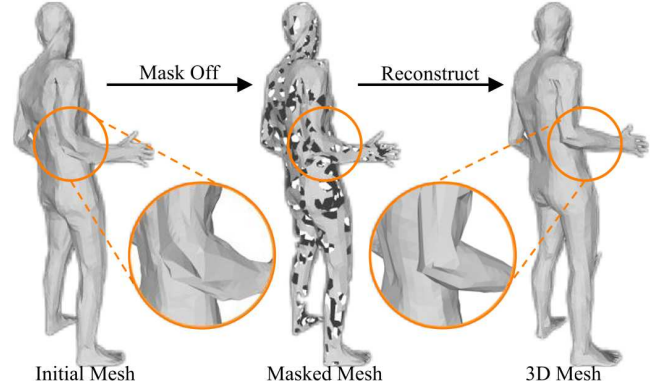}
\caption{Illustration of the proposed shape completion task. With an initial mesh as input, we fabricate artificial holes on the surface of mesh. In order to recover the missing information, the network is forced to reason from the neighborhood in a more distant region. Moreover, we highlight the same part during different phases to show the effectiveness of this module.}
\label{pic:shape}
\end{figure}
By training the network with Eq.~\eqref{eq:Ngcn}, we can obtain a learnable adjacency matrix. However, such stacking of the convolution operation with adaptive adjacency matrix requires the expensive training computational cost. Therefore, we design a classical hierarchical U-shaped network architecture including encoder and decoder parts, to simplify the calculations and achieve our shape reconstruction pipeline.

\begin{table}[!t]
\caption{Comparison on Human3.6M~(Protocol 1 and 2) of our proposed approach with different components~(\ie, adaptive adjacency matrix and U-shaped Net, denoted as A and U separately). The numbers are MPJEP and mean reconstruct errors in mm. We conduct experiments with several models using CMR~\cite{kolotouros2019convolutional} and HMR~\cite{kanazawa2018end-to-end} as the pretrained feature extractors. Best results are in bold.}
\centering
\begin{tabular}{ccccccccccccccc}
\hline
\multicolumn{3}{c|}{\multirow{2}{*}{Method}} & \multicolumn{6}{c|}{MPJPE}                    & \multicolumn{6}{c}{Rec.Error} \\ 
\cline{4-15} 
\multicolumn{3}{c|}{}                        & \multicolumn{3}{c}{P1} & \multicolumn{3}{c|}{P2} & \multicolumn{3}{c}{P1} & \multicolumn{3}{c}{P2} \\ \hline
\multicolumn{3}{l|}{CMR~\cite{kolotouros2019convolutional}}                        &  \multicolumn{3}{c}{77.3}      & \multicolumn{3}{c|}{73.5}              & \multicolumn{3}{c}{51.2}      & \multicolumn{3}{c}{49.6}  \\
\multicolumn{3}{l|}{Ours~(U)}                        &  \multicolumn{3}{c}{{74.7}}      & \multicolumn{3}{c|}{71.0}              & \multicolumn{3}{c}{49.0}      & \multicolumn{3}{c}{46.5}  \\
\multicolumn{3}{l|}{Ours~(A)}                        &  \multicolumn{3}{c}{{73.4}}      & \multicolumn{3}{c|}{69.8}              & \multicolumn{3}{c}{49.1}      & \multicolumn{3}{c}{45.5}  \\
\multicolumn{3}{l|}{Ours~(A+U)}                        &  \multicolumn{3}{c}{\textbf{72.3}}      & \multicolumn{3}{c|}{\textbf{68.8}}              & \multicolumn{3}{c}{\textbf{48.5}}      & \multicolumn{3}{c}{\textbf{45.3}}  \\\hline
\multicolumn{3}{l|}{HMR~\cite{kanazawa2018end-to-end}}                        &  \multicolumn{3}{c}{91.2}      & \multicolumn{3}{c|}{89.1}              & \multicolumn{3}{c}{61.8}      & \multicolumn{3}{c}{59.5}  \\
\multicolumn{3}{l|}{Ours~(U)}                        &  \multicolumn{3}{c}{{89.2}}      & \multicolumn{3}{c|}{87.2}              & \multicolumn{3}{c}{57.9}      & \multicolumn{3}{c}{55.5}  \\
\multicolumn{3}{l|}{Ours~(A)}                        &  \multicolumn{3}{c}{{88.2}}      & \multicolumn{3}{c|}{85.3}              & \multicolumn{3}{c}{56.3}      & \multicolumn{3}{c}{54.1}  \\

\multicolumn{3}{l|}{Ours~(A+U)}                        &  \multicolumn{3}{c}{\textbf{87.8}}      & \multicolumn{3}{c|}{\textbf{85.1}}              & \multicolumn{3}{c}{\textbf{55.6}}      & \multicolumn{3}{c}{\textbf{53.9}} \\
\hline
\end{tabular}
\label{table4Component}
\end{table}
In the encoder part, we introduce sampling operations~\cite{GenerateFaceEccv18} to simplify the calculations. The process of the encoder part can be formulated as: 
\begin{flalign}
{\bm{Y}}_{{l}} = f(\bm{Y}_{l-1}),
\label{eq:pool}
\end{flalign}
where ${\bm{Y}_{l}} \in \mathbb{R}^{\tilde{N} \times {h}}$ is the processed feature matrix with $\tilde{N}$ nodes, and $f$ demonstrates a fully-connected layer. $l \in \{1,2,...l-1,l,l+1,...,L-1,L\}$ is the $l$-th layer of the network. Indeed, by downsampling the mesh data with a pre-defined factor, the high redundancy in the original scale and memory requirements are both dramatically reduced.

In the decoder part, we combine the features to boost the understanding of human body in a coherent way. Similarly, we can denote the decoder part as:
\begin{flalign}
{\bm{Y}}_{{l+1}} = f([f(\bm{Y}_{l});f(m(Y_1,...,Y_l));Y_{L-l}]),
\label{eq:pool_decoder}
\end{flalign}
where $Y_{l+1}$ is a linear combination of above three parts, $m(Y_1,...,Y_l) = \bigcup_{l=1}^l{\sigma(\bm{E}_l\bm{Y}_l\bm{W}_{l})}$ denotes the feature obtained from our 
feature fusion module, and $Y_{L-l}$ is the previous feature in each level of the encoder part in symmetrical module of the decoder part. More specifically, $\bigcup$ represents concatenation connection, $\bm{E}_l \in \mathbb{R}^{\tilde{N} \times \tilde{N}}$ is the calculated attention coefficients matrix and $\bm{W}_{l}\in \mathbb{R}^{h \times p}$ is a shared linear transformation towards $p$ features for each node, which are detailed described in~\cite{velickovic2018graph}.

Essentially, the modeling of the decoder part formulated by Eq.~\eqref{eq:pool_decoder} explicitly fuse multi-level topology information, which alleviates the semantic gap and different spatial resolution~\cite{Qin_2020_PR}. In the process of the decoder part, the body shape is gradually refined when more features are fused, as visualized in Figure~\ref{pic:Process}.

\begin{table}[!t]
\caption{Comparison on MPI-INF-3DHP of our proposed Shape Completion task with different configurations. The numbers are PCK, AUC, and MPJEP in mm. We conduct experiments with CMR~\cite{kolotouros2019convolutional} as pretrained feature extractor. We report results with the different number of nodes that are masked off. Best results are in bold.}
\centering
\begin{tabular}{ccccccccccccccc}
\hline
\multicolumn{3}{c|}{\multirow{2}{*}{Method}} & \multicolumn{12}{c}{Absolute} \\ 
\cline{4-15} 
\multicolumn{3}{c}{}                        & \multicolumn{4}{|l}{PCK~$\uparrow$} & \multicolumn{4}{l}{AUC~$\uparrow$} & \multicolumn{4}{l}{MPJPE~$\downarrow$}  \\ \hline
\multicolumn{3}{l|}{w/o Shape Completion}         &
\multicolumn{4}{l}{62.2} & 
\multicolumn{4}{l}{25.0} & 
\multicolumn{4}{l}{136.2}  \\ 
\multicolumn{3}{l|}{Mask off - 50}                        &  
\multicolumn{4}{l}{63.3} & 
\multicolumn{4}{l}{25.6} & 
\multicolumn{4}{l}{134.6}  \\ 
\multicolumn{3}{l|}{Mask off - 100}                        & 
\multicolumn{4}{l}{64.0} & 
\multicolumn{4}{l}{27.9} & 
\multicolumn{4}{l}{{131.3}}  \\ 
\multicolumn{3}{l|}{Mask off - 200}                        &  
\multicolumn{4}{l}{\textbf{66.3}} & 
\multicolumn{4}{l}{\textbf{30.4}} & 
\multicolumn{4}{l}{\textbf{128.5}}  \\ 
\multicolumn{3}{l|}{Mask off - 400}                        & \multicolumn{4}{l}{64.6} & 
\multicolumn{4}{l}{28.7} & 
\multicolumn{4}{l}{129.7}  \\ 
\hline
\end{tabular}
\label{table4shape}
\end{table}
\subsection{Shape Completion Task}
\label{subsec:shape}

Due to various complex occlusions (\eg,self-occlusion or be sheltered) in an in-the-wild scenario, the human body part information is often missing. Since the training data are collected from simple human actions in a clear indoor environment, the information missing issue is rare. In order to enable the network to learn a generic adjacency matrix for various occlusion cases, we propose a shape completion task in which a mask off and a reconstruction part are included, as shown in Figure~\ref{pic:shape}.

In the mask off part, we simulate the occlusion cases by fabricating artificial holes on the surface of the initial human mesh. 
Specifically, we randomly mask the partial mesh information of a given full human mesh, which can be formulated as 
\begin{flalign}
\hat{\bm{X}}_{in} = \bm{X}_{in}\cdot\bm{M}, 
\label{eq:mask}
\end{flalign}
where $\hat{\bm{X}}_{in}$ denotes a masked human mesh. $\bm{M} \in \mathbb{R}^{N \times {k}}$ is a matrix of ones except $c$ row set zeros, and randomly shuffled before dot product. 

Then, to force the network to recover the missing information for the masked human mesh, 

we replace the $\bm{X}_{in}$ in Eq.~\eqref{eq:Ngcn} as a marked human mesh $\hat{\bm{X}}_{in}$, 
\begin{flalign}
\bm{X}_{in} = \sigma(\hat{\bm{A}}\hat{\bm{X}}_{in}\bm{W}).
\label{eq:gcn_shapeCom}
\end{flalign}

Note that in Eq.~\eqref{eq:gcn_shapeCom}, the output of the network is settled as the initial full human mesh $\bm{X}_{in}$. 
\subsection{Loss Functions}
\label{app:loss}

We use three loss functions to train our network. We first calculate per-vertex $L_1$ loss between the estimated and ground truth shape, which is denoted as $\mathcal{L}_{vertex}$. Additionally, we include joint-wise loss for further aligning mesh with keypoints. Specifically, we implement $L_1$ losses between the projected coordinates and the ground truth keypoints in 2D and 3D space~($J_{2D}$ and $J_{3D}$). Finally, the complete training objective is: 
\begin{flalign}
\mathcal{L}=\mathcal{L}_{vertex}+\mathcal{L}_{3d} +\mathcal{L}_{2d}
\label{loss:all}
\end{flalign}

We provide a more detailed description of the loss function in the supplementary material.

\begin{figure*}[!t]
\centering
\includegraphics[width=\linewidth]{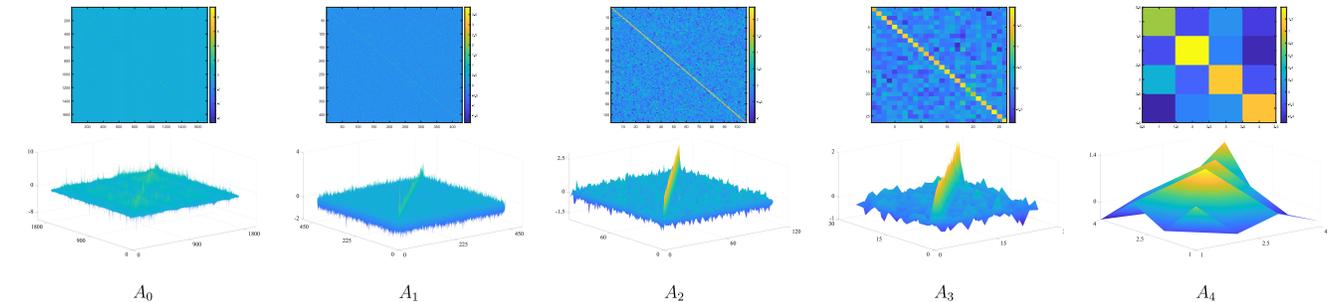}
\caption{Examples of the learned adjacency matrix. For each column, the images from top-to-bottom correspond to the visualization of the learned matrix, the surface plot of the matrix.}
\label{pic:allAdj}
\end{figure*}

\begin{table*}[!t]
\caption{Comparison with state-of-the-art models on MPI-INF-3DHP and Human3.6M datasets~(P2). The numbers are MPJEP and mean reconstruct errors in mm, and AUC. DC-GNet achieves a comparable result on Human3.6M dataset and beyond all state-of-the-art approaches on more challenging in-the-wild MPI-INF-3DHP dataset. ``-" means the corresponding results are not available. $\dagger$ indicates that extra temporal infromation is leveraged. Best results are in bold.}
\centering
\setlength{\tabcolsep}{4mm}{
\begin{tabular}{ccccccccccccccccc}
\hline
\multicolumn{3}{c|}{\multirow{2}{*}{Method}} & \multicolumn{8}{c|}{MPI-INF-3DHP}                               & \multicolumn{6}{c}{Human3.6M}                                \\ 
\cline{4-17} 
\multicolumn{3}{c|}{}                        &\multicolumn{2}{l}{AUC~$\uparrow$}& \multicolumn{3}{l}{MPJPE~$\downarrow$} & \multicolumn{3}{l|}{Reconst.Error~$\downarrow$} & \multicolumn{3}{l}{MPJPE~$\downarrow$} & \multicolumn{3}{l}{Reconst.Error~$\downarrow$} \\ \hline
\multicolumn{3}{l|}{HMR~\cite{kanazawa2018end-to-end}~(CVPR'18)}                       
  & \multicolumn{2}{c}{36.5}& \multicolumn{3}{c}{124.2}      & \multicolumn{3}{c|}{89.8}              & \multicolumn{3}{c}{-}      & \multicolumn{3}{c}{56.8}  \\
  \multicolumn{3}{l|}{$\dagger$HMMR~\cite{KanazawaVideo}~(CVPR'19)}            
  & \multicolumn{2}{c}{-}& \multicolumn{3}{c}{-}      & \multicolumn{3}{c|}{-}              & \multicolumn{3}{c}{-}      & \multicolumn{3}{c}{56.9}  \\
\multicolumn{3}{l|}{$\dagger$Arnab~\etal~\cite{arnab2019exploiting}~(CVPR'19)}    
  & \multicolumn{2}{c}{-}& \multicolumn{3}{c}{-}      & \multicolumn{3}{c|}{-}              & \multicolumn{3}{c}{77.8}      & \multicolumn{3}{c}{54.3}  \\
\multicolumn{3}{l|}{CMR~\cite{kolotouros2019convolutional}~(CVPR'19)}       
  & \multicolumn{2}{c}{24.3}& \multicolumn{3}{c}{152.0}      & \multicolumn{3}{c|}{83.8}              & \multicolumn{3}{c}{71.9}      & \multicolumn{3}{c}{50.1}  \\
\multicolumn{3}{l|}{$\dagger$TexturePose~\cite{pavlakos2019texturepose}~(ICCV'19)} 
  & \multicolumn{2}{c}{-}& \multicolumn{3}{c}{-}      & \multicolumn{3}{c|}{-}              & \multicolumn{3}{c}{-}      & \multicolumn{3}{c}{49.7}  \\
\multicolumn{3}{l|}{SPIN~\cite{kolotouros2019spin}~(ICCV'19)}                   
  & \multicolumn{2}{c}{37.1}& \multicolumn{3}{c}{105.2}      & \multicolumn{3}{c|}{67.5}              & \multicolumn{3}{c}{-}      & \multicolumn{3}{c}{41.1}  \\
\multicolumn{3}{l|}{DaNet~\cite{acmmm19_danet}~(ACM MM'19)}                   
  & \multicolumn{2}{c}{-}& \multicolumn{3}{c}{-}      & \multicolumn{3}{c|}{-}              & \multicolumn{3}{c}{61.5}      & \multicolumn{3}{c}{48.6}  \\
\multicolumn{3}{l|}{Jiang~\etal~\cite{jiang2020coherent}~(CVPR'20)}               
  & \multicolumn{2}{c}{-}& \multicolumn{3}{c}{-}      & \multicolumn{3}{c|}{-}              & \multicolumn{3}{c}{-}      & \multicolumn{3}{c}{52.7}  \\
\multicolumn{3}{l|}{Kundu~\etal~\cite{kundu_human_mesh}~(ECCV'20)}         
  & \multicolumn{2}{c}{-}& \multicolumn{3}{c}{-}      & \multicolumn{3}{c|}{-}              & \multicolumn{3}{c}{-}      & \multicolumn{3}{c}{48.1}  \\
\multicolumn{3}{l|}{Pose2Mesh~\cite{Choi_2020_ECCV_Pose2Mesh}~(ECCV'20)}         
  & \multicolumn{2}{c}{-}& \multicolumn{3}{c}{-}      & \multicolumn{3}{c|}{-}              & \multicolumn{3}{c}{64.9}      & \multicolumn{3}{c}{47.0}  \\
\multicolumn{3}{l|}{$\dagger$VIBE~\cite{kocabas2019vibe}~(CVPR'20)}                        & \multicolumn{2}{c}{-}  & \multicolumn{3}{c}{97.7}      & \multicolumn{3}{c|}{63.4}              & \multicolumn{3}{c}{{65.9}}      & \multicolumn{3}{c}{{41.5}}  \\
\multicolumn{3}{l|}{DecoMR~\cite{DenseCorrespondence}~(CVPR'20)}                        & \multicolumn{2}{c}{-}  & \multicolumn{3}{c}{102.0}      & \multicolumn{3}{c|}{65.9}              & \multicolumn{3}{c}{\textbf{60.6}}      & \multicolumn{3}{c}{\textbf{39.3}}  \\
\hline
\multicolumn{3}{l|}{DC-GNet}                        & \multicolumn{2}{c}{\textbf{40.7}}  & \multicolumn{3}{c}{\textbf{97.2}}      & \multicolumn{3}{c|}{\textbf{62.5}}              & \multicolumn{3}{c}{63.9}      & \multicolumn{3}{c}{42.4}  \\
\hline
\end{tabular}}
\label{table4All}
\end{table*}
\section{Experiment} 
In this Section, we concern with the experimental analysis of the proposed approach. First, we present the datasets that we use for evaluation~(Section~\ref{subsec:Dataset}) and the implementation details of the proposed pipeline~(Section~\ref{subsec:implement}). Then, we discuss the comparison approaches~(Section~\ref{subsec:comApp}) and ablation studies~(Section~\ref{subsec:ablativeS}). Finally, comparison with the-state-of-the-art approaches~(Section~\ref{subsec:com2SOTA}) and qualitative analysis~(Section~\ref{subsec:qualitative}) are provided.

\subsection{Datasets and Evaluation Metrics}
\label{subsec:Dataset}

\textbf{Datasets.} In this paper, we present extensive experiments of our approach on several standard benchmarks including Human3.6M~\cite{2014Human3}, UP-3D~\cite{lassner2017unite}, MPI-INF-3DHP~\cite{mono-3dhp2017}, COCO~\cite{lin2014microsoft} and LSP~\cite{2010Clustered}. For training, we apply benchmarks with 3D annotations, including Human3.6M, UP-3D and MPI-INF-3DHP. Additionally, similar to \cite{kolotouros2019spin}, we incorporate other 2D datasets, \ie, COCO and LSP. For evaluation, we use MPI-INF-3DHP and Human3.6M. For the evaluation on Human3.6M, two popular evaluation protocols can be found. The first one, denoted as P1, includes the subjects S1, S5, S6, S7 and S8 for training, and the subjects S9 and S11 for testing. The second protocol, denoted as P2, tests only on the frontal camera with the same train/test sets. A more detailed description of the datasets can be found in the supplementary material.

\textbf{Evaluation Metrics.} For the MPI-INF-3DHP and Human3.6M datasets, following the evaluation in the most approaches~\cite{Choi_2020_ECCV_Pose2Mesh,kanazawa2018end-to-end,DenseCorrespondence}, we report Mean Per Joint Position Error (MPJPE) and mean reconstruct error. MPJPE is defined as 3D joint errors, which are the projected coordinates from mesh data. While mean reconstruct error is the same calculation with MPJPE but with a rigid alignment. For MPI-INF-3DHP, in addition to MPJPE and mean reconstruct error, we further report Area Under the Curve~(AUC) over a range of Percentage of Correct Keypoints~(PCK) thresholds~\cite{mono-3dhp2017}, which are also used in many approaches\cite{DeeplyComModeccv2018,kanazawa2018end-to-end,kolotouros2019spin}.

\begin{figure*}[]
\centering
\includegraphics[width=0.9\linewidth]{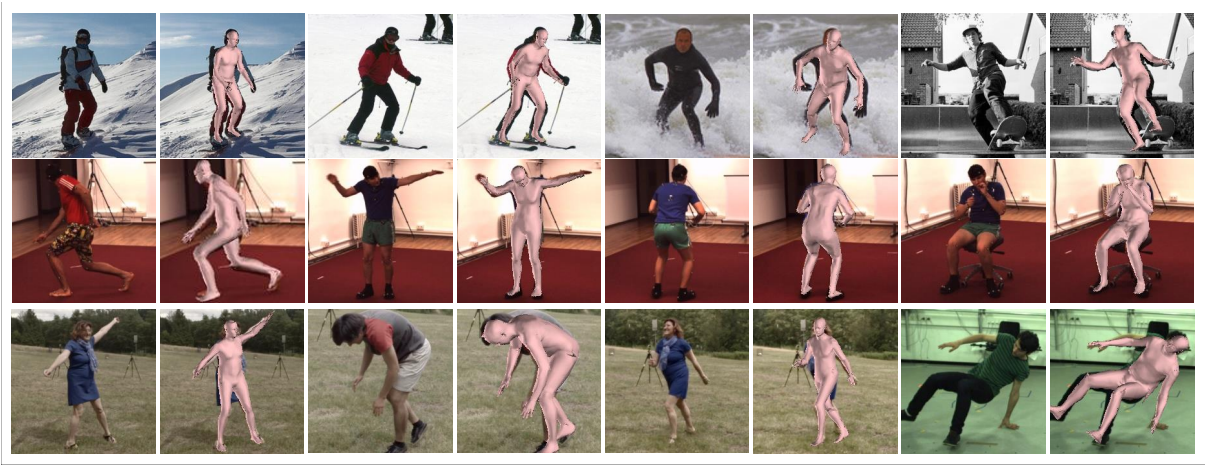}
\caption{Examples of successful reconstructions. COCO~(row 1), H36M~(row 2) and MPI-INF-3DHP~(row 3).}
\label{pic:success}
\end{figure*}

\begin{figure}[]
\centering
\includegraphics[width=\linewidth]{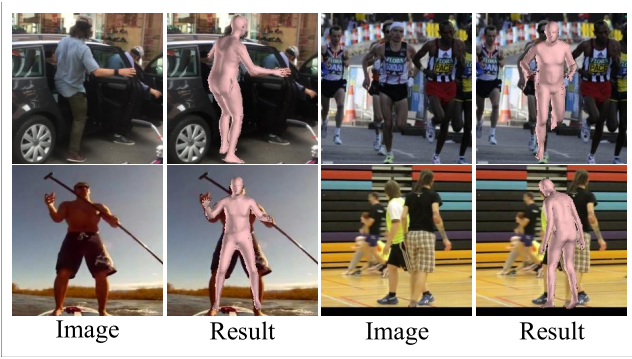}
\caption{Examples of erroneous reconstructions. Typical failure cases may be caused by severe occlusions, rare viewpoint, or interactions among multiple people.}
\label{pic:fail}
\end{figure}
\subsection{Implementation Details}
\label{subsec:implement}
Our model is implemented with PyTorch~\cite{paszke2017automatic}. As shown in Figure~\ref{pic:overview}, we first train our network with a shape completion task, and then in the second step we train the model in an end-to-end fashion. Noted the network trained with shape completion task shares the parameters with the model in the main inference process. In each stage, following \cite{kolotouros2019convolutional} we subsample original mesh by a factor of 4 and upsample it back at the end of the network with \cite{GenerateFaceEccv18}. For the training process, we utilize Adam optimizer with a mini-batch size of 16, where the learning set is set to 3e-4. In the pre-trained process, only Human3.6M dataset is used, while in the main inference process, we first train our model from Human3.6M and UP-3D for 30 epochs and then impose more data~(\ie, COCO, MPI-INF-3DHP, \etc) for greater image diversity. We use a single NVIDIA RTX 2080 Ti GPU for training and our model inference for a single image takes 55ms~(including time~(33ms) for feature extractor), which is nearly real-time.

\subsection{Comparison Approaches}
\label{subsec:comApp}
Following the common-used comparison setting in the literature~\cite{Choi_2020_ECCV_Pose2Mesh,kocabas2019vibe,DenseCorrespondence}, we first compare with two recent baselines for regression-based approaches~(the vertex-level regression and parameter regression approaches, \ie~HMR~\cite{kanazawa2018end-to-end} and CMR~\cite{kolotouros2019convolutional}). As mentioned earlier, both the above two approaches are based on a pre-defined adjacency matrix. Moreover, several of recent state-of-the-art methods~\cite{KanazawaVideo,arnab2019exploiting,pavlakos2019texturepose,kolotouros2019spin,jiang2020coherent,kundu_human_mesh,acmmm19_danet,kolotouros2019convolutional,DenseCorrespondence}, are considered in the comparison, including vertex-level regression approaches~\cite{kolotouros2019convolutional,DenseCorrespondence} and parameter regression ones
\cite{kanazawa2018end-to-end,kolotouros2019spin}.

\subsection{Ablation Studies}
\label{subsec:ablativeS}
Firstly, to put our approach into perspective, we conduct ablation studies on our approach. Following the literature~\cite{kocabas2019vibe}, we also use two pre-trained feature extractors CMR~\cite{kolotouros2019convolutional} and HMR~\cite{kanazawa2018end-to-end}, respectively. We construct three different settings: (a) U-shaped network without adjacency matrix. (b) Only adaptive adjacency matrix (without sampling and features fusion operations). (c) U-shaped network with an adaptive adjacency matrix. These three settings are denoted as "U", "A", and "U+A" in the comparison. The ablation experiments results are reported in Table~\ref{table4Component}, in which the results are organized into two groups in terms of two different feature extractors.

Since the adaptive adjacency matrix captures the topological structure of human body. It significantly improves the reconstruction performance, as we observed in the Table~\ref{table4Component}. The proposed U-shaped network also boosts the performance due to the better understanding of human body in a coherent way. The best result is always achieved by ``U+A" setting whichever feature extractor is used. Figure~\ref{pic:allAdj} visualizes the learned adjacency matrix. It shows that deep relations between nodes (\ie,~positive and negative) are encoded.

\begin{figure*}[]
\centering
\includegraphics[width=0.8\linewidth]{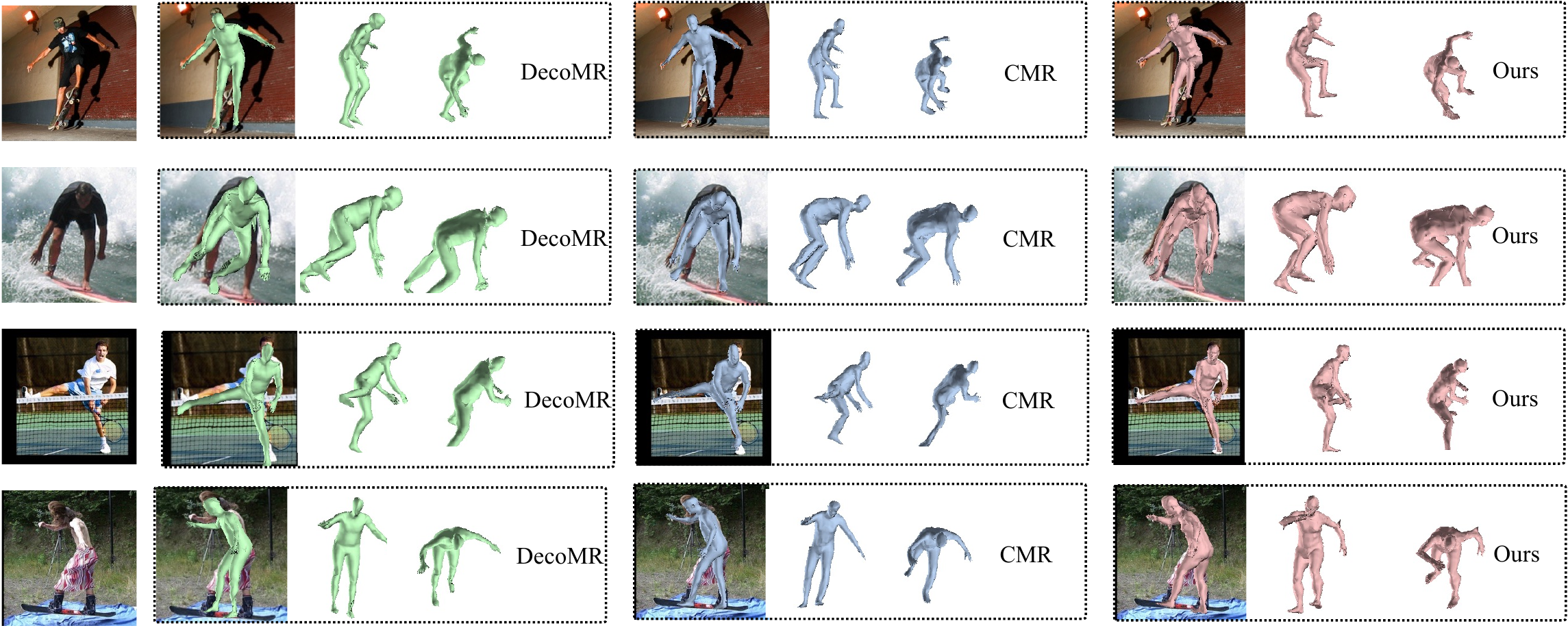}
\caption{Comparison between our approach and other vertex-level regression methods.} 
\label{pic:comp_free}
\end{figure*}
\begin{figure*}[]
\centering
\includegraphics[width=\linewidth]{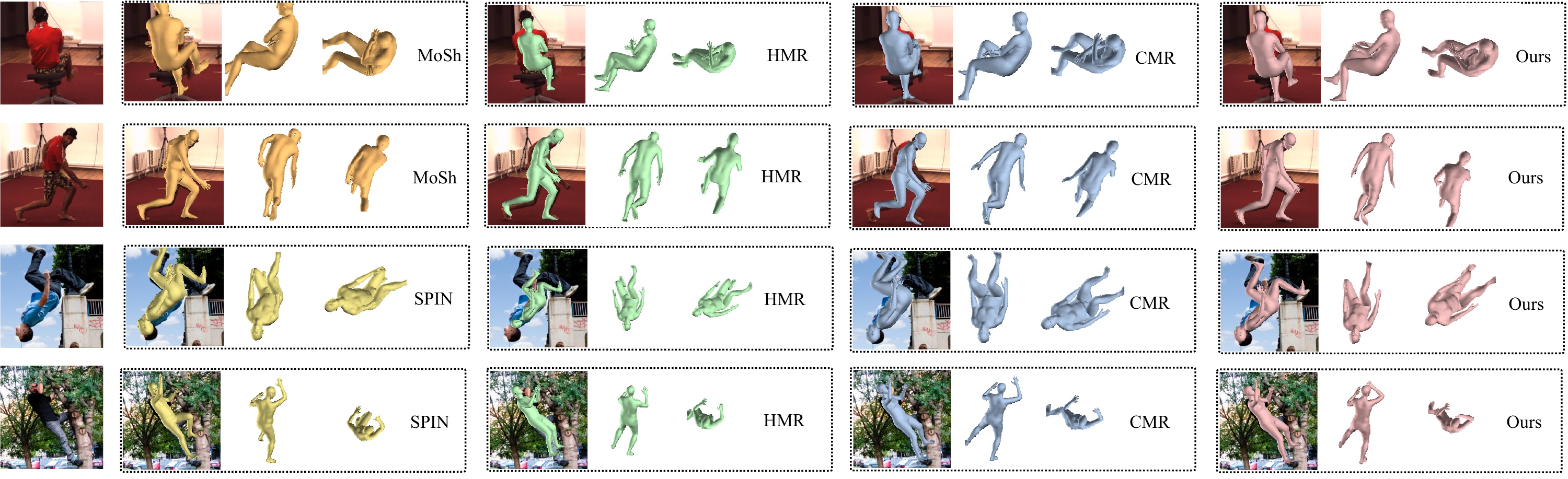}
\caption{Comparison between our approach and other parameter regression methods.
}
\label{pic:MLPCompare}
\end{figure*}
Moreover, we also study the effectiveness of the proposed shape completion task. In the experiment, we train different networks with different number of nodes that are masked off on Human3.6M and UP-3D datasets, and evaluate on MPI-INF-3DHP. The results are obtained by using  CMR~\cite{kolotouros2019convolutional} as a feature extractor. Since the different number of masked nodes leads to different performance, we set the number of masked nodes as 0, 50, 100, 200, 400. As we clearly observed in Table~\ref{table4shape}, with the number of masked nodes increases from 0 to 200, the performance also begins to increase, which demonstrates the effectiveness of the shape completion task. We observe degradation occurs when the number of masked nodes is settled at 400, which may exceed the ability to recover mesh.

\subsection{Comparison to State-of-the-Art Results}
\label{subsec:com2SOTA}
We report MPJPE and mean reconstruct error of DC-GNet on Human3.6M, and additionally AUC over a range of 3D-PCK thresholds~(150mm) on MPI-INF-3DHP. For the fair comparison, following the same setting in the literature~\cite{kocabas2019vibe}, we use HMR as a feature extractor pretrained by~\cite{kolotouros2019spin}.

We first conduct a comparable result on the challenging in-the-wild MPI-INF-3DHP dataset. As shown in Table~\ref{table4All}, comparing to the baseline methods, DC-GNet achieves more than 21\%~(HMR) and 36\%~(CMR) improvement on average MPJPE, respectively. Similarly, more than 30\%~(HMR) and 25\%~(CMR) improvements are achieved by ours on average reconstruct error. Compared to other considered approaches, we still achieve the best performance in all metrics. It seems that our model only achieves sightly performance improvement than \cite{kocabas2019vibe} under MPJPE, note that it exploits video temporal information while we use only a single image.

In the Human3.6M dataset, DC-GNet still shows its superiority. As shown in Table~\ref{table4All}, DC-GNet outperforms the baseline approaches with a wide margin (more than 25\%~(HMR) and 15\%~(CMR) improvement on reconstruction error metric). It seems that our approach is sightly inferior to approaches\cite{DenseCorrespondence,kolotouros2019spin,kocabas2019vibe}. Note that Human3.6M is an indoor dataset with pre-defined action categories in both train and test sets. As analyzed by \cite{Choi_2020_ECCV_Pose2Mesh}, the performance drop (\ie,~performs well on Human3.6M while meets degradation on the in-the-wild dataset) may be attributed to an overfitting issue. 

Note that different methods are trained with different training data. Detailed training data comparisons can be found in the supplementary material. To show the superiority of our approach, we compare their best results reported in the original literature.

\subsection{Qualitative Evaluation}
\label{subsec:qualitative}
This section presents qualitative evaluations. In the qualitative experiments, following the same strategy in~\cite{kolotouros2019convolutional,kanazawa2018end-to-end,kolotouros2019spin}, we leverage Mosh~\cite{mosh2014} and SPIN~\cite{kolotouros2019spin} to generate pseudo-groundtruth on Human3.6M and in-the-wild datasets, respectively.

Firstly, we conduct the qualitative results of our approach from different datasets. The success and failure cases are also reported, as shown in Figure~\ref{pic:success} and Figure~\ref{pic:fail}, respectively. Typical failure cases may be caused by severe occlusions, rare view-point, or interactions among multiple people. 

Moreover, we further provide a qualitative comparison with the recent competitive vertex-level regression approaches~(\ie, CMR~\cite{kolotouros2019convolutional} and DecoMR~\cite{DenseCorrespondence}) and parameter regression approaches~\cite{kanazawa2018end-to-end,kolotouros2019spin}, as shown in Figure~\ref{pic:comp_free} and Figure~\ref{pic:MLPCompare}, respectively. 
As shown in the Figure~\ref{pic:comp_free}, compared to vertex-level regression approaches, DC-GNet generates much more pleasant mesh results that reconstruct details and retain the whole topological structure.

We further report qualitative comparisons with the parameter representation, as shown in Figure~\ref{pic:MLPCompare}. We also achieve more reasonable reconstruction results. Note that CMR and our approach are vertex-level regression approaches while they can also be implemented as a parameter regression by using Multi-Layer Perceptron~(MLP)~\cite{MLP}.

More qualitative results can be found in the supplementary material.
\section{Conclusion}

The aim of this paper is to explore deep relation among mesh vertices for 3D human pose and shape reconstruction, by encoding both positive and negative relations. We incorporate these relations into a graph convolution network with a shape completion module for complex topological structure learning cross domain. 
Moreover, the comparison with a series of state-of-the-art approaches shows the superiority of our approach. More specifically, the extensive experiments conducted on wild datasets demonstrate that the proposed strategies are crucial to make our approach of practical use for in-the-wild scene. Future work may explore using denser cues~(\eg, video input or optical flow) and consider extending our approach for multiple people.

\begin{acks}
  This research was supported by the National Natural Science Foundation of China (Grant no 61671397).
\end{acks}

\bibliographystyle{ACM-Reference-Format}
\bibliography{ref}

\end{document}